\documentclass[10pt,twocolumn,letterpaper]{article}

\usepackage{iccv}
\usepackage{times}
\usepackage{epsfig}
\usepackage{graphicx}
\usepackage{amsmath}
\usepackage{amssymb}
\usepackage[pagebackref=true,breaklinks=true,letterpaper=true,colorlinks,bookmarks=false]{hyperref}
\usepackage{comment}
\usepackage{color}
\usepackage{booktabs}
\usepackage{multirow}
\usepackage{multicol}
\usepackage{xspace}
\usepackage{xcolor}
\usepackage{adjustbox}
\usepackage{import}
\usepackage{subcaption}

\iccvfinalcopy % *** Uncomment this line for the final submission

\def\iccvPaperID{2112} % *** Enter the ICCV Paper ID here

\def\clstoken{\texttt{CLS}\xspace}

\def\ours{CrossViT\xspace}

\newcommand{\boldblue}[1]{{\color{blue}{\textbf{#1}}}}
\newcommand{\myparagraph}[1]{\vspace{1mm} \noindent {\textbf{#1}}}
\newcommand{\myparagraphfirst}[1]{\vspace{0mm} \noindent {\textbf{#1}}}

\ificcvfinal\fi

\graphicspath{{Main/},{Supplemental/}}

\begin{document}

%%%%%%%%% TITLE
\title{CrossViT: Cross-Attention Multi-Scale Vision Transformer for Image Classification}

\author{Chun-Fu (Richard) Chen, Quanfu Fan, Rameswar Panda\\
MIT-IBM Watson AI Lab\\
% Institution1 address\\
{\tt\small chenrich@us.ibm.com, qfan@us.ibm.com, rpanda@ibm.com}
% For a paper whose authors are all at the same institution,
% omit the following lines up until the closing ``}''.
% Additional authors and addresses can be added with ``\and'',
% just like the second author.
% To save space, use either the email address or home page, not both
% \and
% Second Author\\
% Institution2\\
% First line of institution2 address\\
% {\tt\small secondauthor@i2.org}
}

\maketitle
% Remove page # from the first page of camera-ready.
\ificcvfinal\thispagestyle{empty}\fi

%%%%%%%%% ABSTRACT
\begin{abstract}
The recently developed vision transformer (ViT) has achieved promising results on image classification compared to convolutional neural networks. Inspired by this, in this paper, we study how to learn multi-scale feature representations in transformer models for image classification. To this end, we propose a dual-branch transformer to combine image patches (i.e., tokens in a transformer) of different sizes to produce stronger image features. Our approach processes small-patch and large-patch tokens with two separate branches of different computational complexity and these tokens are then fused purely by attention multiple times to complement each other. Furthermore, to reduce computation, we develop a simple yet effective token fusion module based on cross attention, which uses a single token for each branch as a query to exchange information with other branches. Our proposed cross-attention only requires linear time for both computational and memory complexity instead of quadratic time otherwise. Extensive experiments demonstrate that our approach performs better than or on par with several concurrent works on vision transformer, in addition to efficient CNN models. For example, on the ImageNet1K dataset, with some architectural changes, our approach outperforms the recent DeiT by a large margin of 2\% with a small to moderate increase in FLOPs and model parameters. Our source codes and models are available at \url{https://github.com/IBM/CrossViT}.

\end{abstract}

\section{Introduction}
\label{sec:intro}

\begin{figure}[tb!]
    \centering
    \includegraphics[width=.9\linewidth]{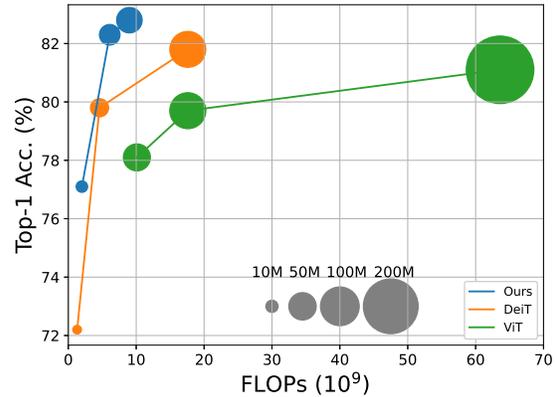}
    \caption{\textbf{Improvement of our proposed approach over DeiT~\cite{DeiT_touvron2020} and ViT~\cite{ViT_dosovitskiy2021an}}. The circle size is proportional to the model size. All models are trained on ImageNet1K from scratch. The results of ViT are referenced from~\cite{tokenstotoken_yuan2021}.
    %\rp{Since we are putting a dagger, just writing Ours may be good for the first figure? We can change it to the previous one}
    }  
    \label{fig:perf_comp} 
    % \vspace{-3mm}
\end{figure}
% \QF{change 'CrossViT+ to 'proposed' or something similar in Fig. 1?} \rc{changed to ours}

% the surging of transformers in vision
The novel transformer architecture~\cite{Transformer_NIPS2017_Vaswani} has led to a big leap forward in capabilities for sequence-to-sequence modeling in NLP tasks~\cite{devlin2018bert}. The great success of transformers in NLP has sparked particular interest from the vision community in understanding whether transformers can be a strong competitor against the dominant Convolutional Neural Network based architectures (CNNs) in vision tasks such as ResNet~\cite{ResNet_He_2016_CVPR} and EfficientNet~\cite{efficientnet_pmlr_tan_19}. Previous research  efforts on transformers in vision have, until very recently,  been largely focused on combining CNNs with self-attention~\cite{bello2019attention, SAN_Zhao_2020_CVPR, SASA_Ramachandran_2019_NeurIPS, BoT_srinivas2021}. While these hybrid approaches achieve promising performance, they have limited scalability in computation compared to purely attention-based transformers. Vision Transformer (ViT)~\cite{ViT_dosovitskiy2021an}, which uses a sequence of embedded image patches as input to a standard transformer, is the first kind of convolution-free transformers that demonstrate comparable performance to CNN models. However, ViT requires very large datasets such as ImageNet21K~\cite{imagenet_deng2009} and JFT300M~\cite{JFT300M_ICCV_2017} for training. DeiT~\cite{DeiT_touvron2020} subsequently shows that data augmentation and model regularization can enable training of high-performance ViT models with fewer data. Since then, ViT has instantly inspired several attempts to improve its efficiency and effectiveness from different aspects~\cite{DeiT_touvron2020, tokenstotoken_yuan2021, TNT_han2021transformer,PVT_wang2021, perceiver_jaegle2021}.

Along the same line of research on building stronger vision transformers, in this work, we study \textit{how to learn multi-scale feature representations in transformer models for image recognition}. Multi-scale feature representations have proven beneficial for many vision tasks~\cite{chen2018big,cai2016unified,lin2017feature,SKNet_Li_2019_CVPR,hourglass,Deblurring_Nah_2017_CVPR,HigherHRNet_Cheng_2020_CVPR}, but such potential benefit for vision transformers remains to be validated. %It is shown in ViT ~\cite{ViT_dosovitskiy2021an} that a transformer model based on smaller patches (i.e. 14x14) are more accurate than a one based on larger patches (i.e. 16x16). This hints that combining image patches of different sizes xxxx. 
Motivated by the effectiveness of multi-branch CNN architectures such as Big-Little Net~\cite{chen2018big} and Octave convolutions~\cite{chen2019drop}, we  propose  a  dual-branch  transformer  to  combine image  patches  (i.e. tokens  in  a  transformer)  of  different sizes  to  produce  stronger visual features for image classification. Our approach processes  small and large patch tokens with two separate  branches of different computational complexities and  these  tokens are fused together multiple times to  complement each other. Our main focus of this work is to develop feature fusion methods that are appropriate for vision transformers, which has not been addressed to the best of our knowledge. We do so by an efficient cross-attention module, in which each transformer branch creates a non-patch token as an agent to exchange information with the other branch by attention.  This allows for linear-time  generation  of  the  attention  map  in fusion instead of quadratic time otherwise. With some proper architectural adjustments in computational loads of each branch, our proposed approach outperforms DeiT~\cite{DeiT_touvron2020} by a large margin of 2\% with a small to moderate increase in FLOPs and model parameters (See Figure~\ref{fig:perf_comp}).
%motivation: why multi-scale transformer?

The main contributions of our work are as follows:
\begin{itemize}
    \item We propose a novel dual-branch vision transformer to extract multi-scale feature representations for image classification. Moreover, we develop a simple yet effective token fusion scheme based on cross-attention, which is linear in both computation and memory to combine features at different scales. 
    %The results show that our models consistently outperform ViT by $\sim$2\% with slightly increase on the complexity (\QF{? a more accurate statement?}). 
    % \rc{How about in the first point, we mention the gain of cross attention only first and then the overall gain? like: The results show that our proposed multi-scale fusion consistently improve ViT by $\sim$1.0\% slightly increase on the complexity; furthermore, with minor modification on the network architecture, our models can surpass ViT by up to X\%. ? Or we would like to always use the best numbers to say that?}
    %\item 
    \item Our approach performs better than or on par with several concurrent works based on ViT~\cite{ViT_dosovitskiy2021an}, and demonstrates comparable results with EfficientNet~\cite{efficientnet_pmlr_tan_19} with regards to accuracy, throughput and model parameters. %We further combine their approaches into our work, and the resulted models achieve extra 0.5$\sim$0.7\% point improvement.
\end{itemize}
    
\section{Related Works}
\label{sec:related_works}

Our work relates to three major research directions: convolutional neural networks with attention, vision transformer and multi-scale CNNs. Here, we focus on some representative methods closely related to our work.

\vspace{0.5mm}
\myparagraphfirst{CNN with Attention.} Attention has been widely used in many different forms to enhance feature representations, e.g., SENet~\cite{SENet_Hu_2018} uses channel-attention, CBAM~\cite{CBAM_Woo_2018_ECCV} adds the spatial attention and ECANet~\cite{ECA_wang2020} proposes an efficient channel attention to further improve SENet. There has also been a lot of interest in combining CNNs with different forms of self-attention~\cite{lambdanetworks_bello2021,BoT_srinivas2021,SAN_Zhao_2020_CVPR,SASA_Ramachandran_2019_NeurIPS,bello2019attention,hu2019local,wang2018non}. SASA~\cite{SASA_Ramachandran_2019_NeurIPS} and SAN~\cite{SAN_Zhao_2020_CVPR} deploy a local-attention layer to replace convolutional layer. Despite promising results, prior approaches limited the attention scope to local region due to its complexity. LambdaNetwork~\cite{lambdanetworks_bello2021} recently introduces an efficient global attention to model both content and position-based interactions that considerably improves the speed-accuracy tradeoff of image classification models.
BoTNet~\cite{BoT_srinivas2021} replaced the spatial convolutions with global self-attention in the final three bottleneck blocks of a ResNet resulting in models that achieve a strong performance for image classification on ImageNet benchmark.
In contrast to these approaches that mix convolution with self-attention, our work is built on top of pure self-attention network like Vision Transformer~\cite{ViT_dosovitskiy2021an} which has recently shown great promise in several vision applications. 

\myparagraph{Vision Transformer.} Inspired by the success of Transformers~\cite{Transformer_NIPS2017_Vaswani} in machine translation, convolution-free models that only rely on transformer layers have gone viral in computer vision. In particular, Vision Transformer (ViT)~\cite{ViT_dosovitskiy2021an} is the first such example of a transformer-based method to match or even surpass CNNs for image classification. Many variants of vision transformers have also been recently proposed that uses distillation for data-efficient training of vision transformer~\cite{DeiT_touvron2020}, pyramid structure like CNNs~\cite{PVT_wang2021}, or self-attention to improve the efficiency via learning an abstract representation instead of performing all-to-all self-attention~\cite{centroidvit_wu2021}. Perceiver~\cite{perceiver_jaegle2021} leverages an asymmetric attention mechanism to iteratively distill inputs into a tight latent bottleneck, allowing it to scale to handle very large inputs. T2T-ViT~\cite{tokenstotoken_yuan2021} introduces a layer-wise Tokens-to-Token (T2T) transformation to encode the important local structure for each token instead of the naive tokenization used in ViT~\cite{ViT_dosovitskiy2021an}.
Unlike these approaches, we propose a dual-path architecture to extract multi-scale features for better visual representation with vision transformers.

%T2T-ViT~\cite{tokenstotoken_yuan2021} proposed the tokens-to-token module to encode an image into tokens. Instead of directly performing linear projection from a patch into a token, they structurized an image progressively by iteratively aggregating neighboring pixels into tokens.
%show the simple architecture with self-attention achieved promising results on image classification as compared to convolutional neural networks when having a huge amount of data. After that, many variants of ViT are proposed by incorporating the knowledge of computer vision field. 
% Perceiver XXXX
%Pyramid vision transformer (PVT)~\cite{PVT_wang2021} uses pyramid structure like CNNs: more patch tokens and smaller channel size in the front layers while fewer patch tokens and wider channel size at the rear layers.
%CentroidViT~\cite{centroidvit_wu2021} proposed an efficient self-attention to improve the efficiency via learning an abstract representation instead of performing all-to-all self-attention. 

\myparagraph{Multi-Scale CNNs.} Multi-scale feature representations have a long  history in computer vision (e.g., image pyramids~\cite{adelson1984pyramid}, scale-space representation~\cite{perona1990scale}, and coarse-to-fine approaches~\cite{pedersoli2015coarse}).
In the context of CNNs, multi-scale feature representations have been used for detection and recognition of objects at multiple scales~\cite{cai2016unified,lin2017feature,yang2015multi,newell2016stacked}, as well as to speed up neural networks in Big-Little Net~\cite{chen2018big} and OctNet~\cite{chen2019drop}. bLVNet-TAM \cite{fan2019more} uses a two-branch multi-resolution architecture while learning temporal dependencies across frames. SlowFast Networks~\cite{feichtenhofer2019slowfast} rely on a similar two-branch model, but each branch encodes different frame rates, as opposed to frames with different spatial resolutions. While multi-scale features have shown to benefit CNNs, it's applicability for vision transformer still remains as a novel and largely under-addressed problem. 

% 

%In the vision domain, an object can has different scales, so extracting multi-scale feature is always an important step for better performance.

\section{Method}
\label{sec:proposed}
Our method is built on top of vision transformer~\cite{ViT_dosovitskiy2021an}, so we first present a brief overview of ViT and then describe our proposed method (\ours) for learning multi-scale features for image classification.

\begin{figure}[tb!]
    \centering
    \includegraphics[width=\linewidth]{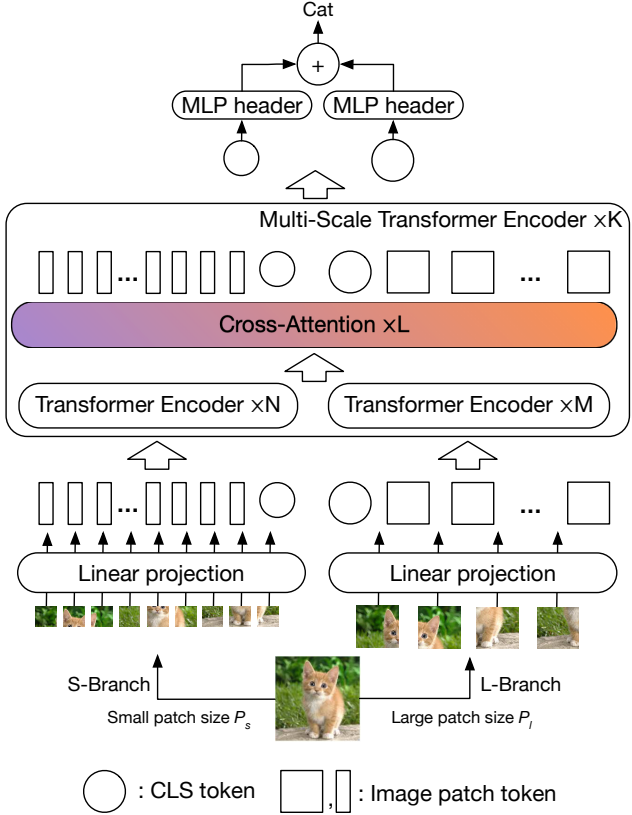}
    \caption{\textbf{An illustration of our proposed transformer architecture for learning multi-scale features with cross-attention (\ours).} Our architecture consists of a stack of $K$ multi-scale transformer encoders. Each multi-scale transformer encoder uses two different branches to process image tokens of different sizes ($P_s$ and $P_l$, $P_s < P_l$) and fuse the tokens at the end by an efficient module based on cross attention of the \clstoken tokens. Our design includes different numbers of regular transformer encoders in the two branches (i.e. N and M) to balance computational costs. 
    }
    \label{fig:cross_attention} 
    % \vspace{-3mm}
\end{figure}

\subsection{Overview of Vision Transformer}

Vision Transformer (ViT)~\cite{ViT_dosovitskiy2021an} first converts an image into a sequence of patch tokens by dividing it with a certain patch size and then linearly projecting each patch into tokens.  An additional classification token (\clstoken) is added to the sequence, as in the original BERT~\cite{devlin2018bert}.
Moreover, since self-attention in the transformer encoder is position-agnostic and vision applications highly need position information, ViT adds position embedding into each token, including the \clstoken token. Afterwards, all tokens are passed through stacked transformer encoders and finally the \clstoken token is used for classification. A transformer encoder is composed of a sequence of blocks where each block contains multiheaded self-attention ($\mathtt{MSA}$) with a feed-forward network ($\mathtt{FFN}$). 
$\mathtt{FFN}$ contains two-layer multilayer perceptron with expanding ratio $r$ at the hidden layer, and one GELU non-linearity is applied after the first linear layer.
Layer normalization ($\mathtt{LN}$) is applied before every block, and residual shortcuts after every block. 
The input of ViT, $\mathbf{x}_0$, and the processing of the $k$-th block can be expressed as
\begin{equation} 
\begin{split}
\label{eq:vit}
    \mathbf{x}_0 & = \left[ \mathbf{x}_{cls} \vert \vert \mathbf{x}_{patch} \right] + \mathbf{x}_{pos} \\
    \mathbf{y}_k & = \mathbf{x}_{k-1} + \mathtt{MSA}(\mathtt{LN}(\mathbf{x}_{k-1})) \\
    \mathbf{x}_k & = \mathbf{y}_k + \mathtt{FFN}(\mathtt{LN}(\mathbf{y}_k)),
\end{split}
\end{equation}

where $\mathbf{x}_{cls} \in \mathbb{R}^{1\times C}$ and $\mathbf{x}_{patch} \in \mathbb{R}^{N\times C}$ are the \clstoken and patch tokens respectively and $\mathbf{x}_{pos} \in \mathbb{R}^{(1+N)\times C}$ is the position embedding. $N$ and $C$ are the number of patch tokens and dimension of the embedding, respectively.

\begin{figure*}[tbh!]
    \centering
    \includegraphics[width=\linewidth]{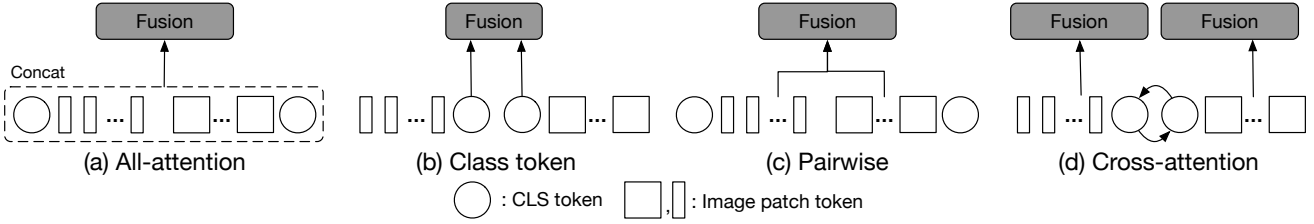}
    % \vspace{-6mm}
    \caption{\textbf{Multi-scale fusion}. 
    (a) All-attention fusion where all tokens are bundled together without considering any characteristic of tokens. 
    (b) Class token fusion, where only \clstoken tokens are fused as it can be considered as global representation of one branch. (c) Pairwise fusion, where tokens at the corresponding spatial locations are fused together and \clstoken are fused separately. (d) Cross-attention, where \clstoken token from one branch and patch tokens from another branch are fused together. 
    }
    \label{fig:fusions} 
    % \vspace{-3mm}
\end{figure*}

It is worth noting that one very different design of ViT from CNNs is the \clstoken token. In CNNs, the final embedding is usually obtained by averaging the features over all spatial locations while ViT uses the \clstoken that interacts with patch tokens at every transformer encoder as the final embedding. Thus, we consider \clstoken as an agent that summarizes all the patch tokens and hence the proposed module is designed based on \clstoken to form a dual-path multi-scale ViT.

\subsection{Proposed Multi-Scale Vision Transformer}
\label{subsec:crossvit_overview}
The granularity of the patch size affects the accuracy and complexity of ViT; with fine-grained patch size, ViT can perform better but results in higher FLOPs and memory consumption. For example, the ViT with a patch size of 16 outperforms the ViT with a  patch size of 32 by 6\% but the former needs 4$\times$ more FLOPs. Motivated by this, our proposed approach is trying to leverage the advantages from more fine-grained patch sizes while balancing the complexity. More specifically, we first introduce a dual-branch ViT where each branch operates at a different scale (or patch size in the patch embedding) and then propose a simple yet effective module to fuse information between the branches.

Figure~\ref{fig:cross_attention} illustrates the network architecture of our proposed Cross-Attention Multi-Scale Vision Transformer (\ours). Our model is primarily composed of $K$ multi-scale transformer encoders where each encoder consists of two branches: (1) \textbf{L-Branch}: a \textit{large (primary)} branch that utilizes coarse-grained patch size ($P_l$) with more transformer encoders and wider embedding dimensions, (2) \textbf{S-Branch}: a \textit{small (complementary)} branch that operates at fine-grained patch size ($P_s$) with fewer encoders and smaller embedding dimensions. Both branches are fused together $L$ times and the \clstoken tokens of the two branches at the end are used for prediction. 
Note that for each token of both branches, we also add a learnable position embedding before the multi-scale transformer encoder for learning position information as in ViT~\cite{ViT_dosovitskiy2021an}.

Effective feature fusion is the key for learning multi-scale feature representations. We explore four different fusion strategies: three simple heuristic approaches and the proposed cross-attention module as shown in Figure~\ref{fig:fusions}. Below we provide the details on these fusion schemes.

\subsection{Multi-Scale Feature Fusion}
\label{subsec:ms_fusion}

Let $\mathbf{x}^i$ be the token sequence (both patch and \clstoken tokens) at branch $i$, where $i$ can be $l$ or $s$ for the large (primary) or small (complementary) branch. 
$\mathbf{x}^i_{cls}$ and $\mathbf{x}^i_{patch}$ represent \clstoken and patch tokens of branch $i$, respectively. 

\vspace{1mm}
\myparagraph{All-Attention Fusion.}
A straightforward approach is to simply concatenate all the tokens from both branches without considering the property of each token and then fuse information via the self-attention module, as shown in Figure~\ref{fig:fusions}(a). 
This approach requires quadratic computation time since all tokens are passed through the self-attention module. 
The output $\mathbf{z}^i$ of the all-attention fusion scheme can be expressed as 
\begin{equation} 
\begin{split}
\label{eq:all_attn}
    \mathbf{y} & = \left[ f^l(\mathbf{x}^l) \ \vert\vert \ f^s(\mathbf{x}^s) \right], \ \  \mathbf{o} = \mathbf{y} + \mathtt{MSA}(\mathtt{LN}(\mathbf{y})), \\ 
    \mathbf{o} & = \left[ \mathbf{o}^l \ \vert\vert \ \mathbf{o}^s \right], \ \ \mathbf{z}^i = g^i(\mathbf{o}^i), \\
\end{split}
\end{equation} 
where $f^i(\cdot)$ and $g^i(\cdot)$ are the projection and back-projection functions to align the dimension.

\vspace{1mm}
\myparagraph{Class Token Fusion.} 
The \clstoken token can be considered as an abstract global feature representation of a branch since it is used as the final embedding for prediction. Thus, a simple approach is to sum the \clstoken tokens of two branches, as shown in Figure~\ref{fig:fusions}(b). This approach is very efficient as only one token needs to be processed. Once \clstoken tokens are fused, the information will be passed back to patch tokens at the later transformer encoder. More formally, the output $\mathbf{z}^i$ of this fusion module can be represented as 
\begin{equation} 
\begin{split}
\label{eq:sum_cls}
\mathbf{z}^i = \left[ g^i(\sum_{j \in \{l, s\}} {f^j(\mathbf{x}_{cls}^j)}) \ \vert\vert \ \mathbf{x}^i_{patch} \right],
\end{split}
\end{equation} 
where $f^i(\cdot)$ and $g^i(\cdot)$ play the same role as Eq.~\ref{eq:all_attn}.

\vspace{1mm}
\myparagraph{Pairwise Fusion.} 
Figure~\ref{fig:fusions}(c) shows how both branches are fused in pairwise fusion. Since patch tokens are located at its own spatial location of an image, a simple heuristic way for fusion is to combine them based on their spatial location. However, the two branches process patches of different sizes, thus having different number of patch tokens. We first perform an interpolation to align the spatial size, and then fuse the patch tokens of both branches in a pair-wise manner. On the other hand, the two \clstoken are fused separately. The output $\mathbf{z}^i$ of pairwise fusion of branch $l$ and $s$ can be expressed as
\begin{equation} 
\begin{split}
\label{eq:sum_all}
    \mathbf{z}^i = \left[ g^i(\sum_{j \in \{l, s\}} {f^j(\mathbf{x}_{cls}^j)}) \ \vert\vert \ g^i(\sum_{j \in \{l, s\}} {f^j(\mathbf{x}_{patch}^j)}) \right],
\end{split}
\end{equation} 
where $f^i(\cdot)$ and $g^i(\cdot)$ play the same role as Eq.~\ref{eq:all_attn}.

\begin{figure}[tb!]
    \centering
    \includegraphics[width=.7\linewidth]{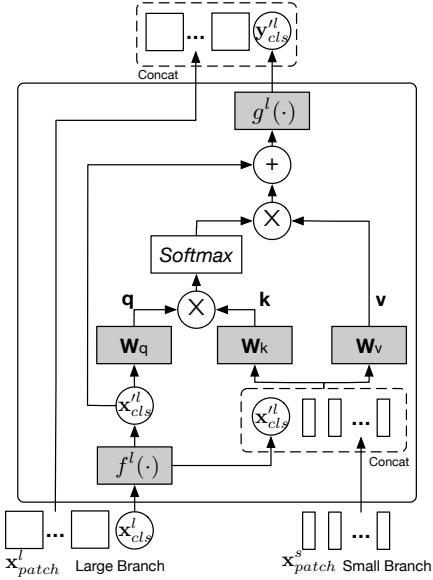}
    \caption{\textbf{Cross-attention module for Large branch}. The \clstoken token of the large branch (circle) serves as a query token to interact with the patch tokens from the small branch through attention.
    $f^l(\cdot)$ and $g^l(\cdot)$ are projections to align dimensions. The small branch follows the same procedure but swaps \clstoken and patch tokens from another branch. 
    }
    \label{fig:ca_module} 
%   \vspace{-3mm}
\end{figure}

\myparagraph{Cross-Attention Fusion.}
Figure~\ref{fig:fusions}(d) shows the basic idea of our proposed cross-attention, where the fusion involves the \clstoken token of one branch and patch tokens of the other branch. 
Specifically, in order to fuse multi-scale features more efficiently and effectively, we first utilize the \clstoken token at each branch as an agent to exchange information among the patch tokens from the other branch and then back project it to its own branch. Since the \clstoken token already learns abstract information among all patch tokens in its own branch, interacting with the patch tokens at the other branch helps to include information at a different scale. After the fusion with other branch tokens, the \clstoken token interacts with its own patch tokens again at the next transformer encoder, where it is able to pass the learned information from the other branch to its own patch tokens, to enrich the representation of each patch token. In the following, we describe the cross-attention module for the large branch (L-branch), and the same procedure is performed for the small branch (S-branch) by simply swapping the index $l$ and $s$. 

An illustration of the cross-attention module for the large branch is shown in Figure~\ref{fig:ca_module}. 
Specifically, for branch $l$, it first collects the patch tokens from the S-Branch and concatenates its own \clstoken tokens to them, as shown in Eq.~\ref{eq:ca_1}.

\begin{equation} 
\begin{split}
\label{eq:ca_1}
    \mathbf{x}'^{l} = \left[f^l(\mathbf{x}^l_{cls}) \ \vert\vert \ \mathbf{x}^{s}_{patch}  \right], \\
\end{split}
\end{equation} 
where $f^l(\cdot)$ is the projection function for dimension alignment.
The module then performs cross-attention ($\mathtt{CA}$) between $\mathbf{x}_{cls}^l$ and $\mathbf{x}'^l$, where \clstoken token is the only query as the information of patch tokens are fused into \clstoken token. Mathematically, the $\mathtt{CA}$ can be expressed as  
\begin{equation} 
\begin{split}
\label{eq:cls}
    \mathbf{q} = {\mathbf{x}'^l_{cls}}\mathbf{W}_q, \ \ \ \mathbf{k} = \mathbf{x}'^l\mathbf{W}_k, \ \ \ \mathbf{v} = \mathbf{x}'^l\mathbf{W}_v, \\
    \mathbf{A} = \mathtt{softmax}(\mathbf{q}\mathbf{k}^T / \sqrt{C / h} ), \ \ \ 
    \mathtt{CA}(\mathbf{x}'^l) = \mathbf{A}\mathbf{v},
\end{split}
\end{equation}
where $\mathbf{W}_q$, $\mathbf{W}_k$, $\mathbf{W}_v$ $\in \mathbb{R}^{C\times (C/h)}$ are learnable parameters, $C$ and $h$ are the embedding dimension and number of heads.
Note that since we only use \clstoken in the query, the computation and memory complexity of generating the attention map ($\mathbf{A}$) in cross-attention are linear rather than quadratic as in all-attention, making the entire process more efficient. Moreover, as in self-attention, we also use multiple heads in the $\mathtt{CA}$ and represent it as ($\mathtt{MCA}$). However, we do not apply a feed-forward network $\mathtt{FFN}$ after the cross-attention. 
Specifically, the output $\mathbf{z}^l$ of a cross-attention module of a given $\mathbf{x}^l$ with layer normalization and residual shortcut is defined as follows.
\begin{equation} 
\begin{split}
\label{eq:cls_transformer}
    \mathbf{y}^l_{cls} = & f^l\left(\mathbf{x}^l_{cls}\right) + \mathtt{MCA}(\mathtt{LN}(\left[f^l(\mathbf{x}^l_{cls}) \ \vert\vert \ \mathbf{x}^{s}_{patch}  \right])) \\
    \mathbf{z}^l = & \left[ g^l\left(\mathbf{y}^l_{cls}\right) \ \vert\vert \ \mathbf{x}^l_{patch}  \right],
\end{split}
\end{equation}
where $f^l(\cdot)$ and $g^l(\cdot)$ are the projection and back-projection function for dimension alignment, respectively.
We empirically show in Section~\ref{subsec:ablations} that cross-attention achieves the best accuracy compared to other three simple heuristic approaches while being efficient for mult-scale feature fusion.

\section{Experiments}
\label{sec:exp}

In this section, we conduct extensive experiments to show the effectiveness of our proposed \ours over existing methods. First, we check the advantages of our proposed model over the baseline DeiT in Table~\ref{table:baseline}, and then we compare with serveral concurrent ViT variants and CNN-based models in Table~\ref{table:sota} and Table~\ref{table:compare_cnn}, respectively. Moreover, we also test the transferability of \ours on 5 downstream tasks (Table~\ref{table:transfer}). Finally, we perform ablation studies on different fusion schemes in Table~\ref{table:ablation_fusion} and discuss the effect of different parameters of \ours in Table~\ref{table:ablation}.

%\subsection{Models, Datasets and Training Details}
\subsection{Experimental Setup}
\myparagraphfirst{Dataset.}
We validate the effectiveness of our proposed approach on the ImageNet1K dataset~\cite{imagenet_deng2009}, and use the top-1 accuracy on the validation set as the metrics to evaluate the performance of a model. ImageNet1K contains 1,000 classes and the number of training and validation images are 1.28 millions and 50,000, respectively. 
We also test the transferability of our approach using several smaller datasets, such as CIFAR10~\cite{cifar_krizhevsky2009learning} and CIFAR100~\cite{cifar_krizhevsky2009learning}. 
% \rc{should we list all? or put etc?}

\myparagraph{Training and Evaluation.}
The original ViT~\cite{ViT_dosovitskiy2021an} achieves competitive results compared to some of the best CNN models but only when trained on very large-scale datasets (e.g. ImageNet21K~\cite{imagenet_deng2009} and JFT300M~\cite{JFT300M_ICCV_2017}). Nevertheless, DeiT~\cite{DeiT_touvron2020} shows that  with the help of a rich set of data augmentation techniques, ViT can be trained from ImageNet alone to produce comparable results to CNN models. Therefore, in our experiments, we build our models based on DeiT~\cite{DeiT_touvron2020}, and apply their default hyper-parameters for training. These data augmentation methods include rand augmentation~\cite{RandAug_NEURIPS2020_Cubuk}, mixup~\cite{Mixup_zhang2018} and cutmix~\cite{CutMix_Yun_2019_ICCV} as well as random erasing~\cite{RandomErasing_Zhong_2020}. We also apply drop path~\cite{efficientnet_pmlr_tan_19} for model regularization but instance repetition~\cite{InstanceRepetition_Hoffer_2020_CVPR} is only enabled for \ours-18 as it does not improve small models. 

We train all our models for 300 epochs (30 warm-up epochs) on 32 GPUs with a batch size of 4,096. Other setup includes a cosine linear-rate scheduler with linear warm-up, an initial learning rate of 0.004 and a weight decay of 0.05. During evaluation, we resize the shorter side of an image to 256 and take the center crop 224$\times$224 as the input.
Moreover, we also fine-tuned our models with a larger resolution (384$\times$384) for fair comparison in some cases. Bicubic interpolation was applied to adjust the size of the learnt position embedding, and the finetuning took 30 epochs. More details can be found in supplementary material.
%and the attached codes.

\myparagraph{Models.}
Table~\ref{table:models} specifies the architectural configurations of the \ours models used in our evaluation. Among these models, \ours-Ti, \ours-S and \ours-B set their large (primary) branches identical to the tiny (DeiT-Ti), small (DeiT-S) and base (DeiT-B) models introduced in DeiT~\cite{DeiT_touvron2020}, respectively. 
%The model with letter suffix (T, S) means the \textit{primary} branch is the corresponding ViT-X (DeiT-X) models~\cite{ViT_dosovitskiy2021an, DeiT_touvron2020}; 
The other models vary by different expanding ratios in $\mathtt{FFN}$ ($r$), depths and embedding dimensions. In particular, the ending number in a model name tells the total number of transformer encoders in the large branch used. For example, \ours-15 has 3 multi-scale encoders, each of which includes 5 regular transformers, resulting in a total of 15 transformer encoders. 

The original ViT paper~\cite{ViT_dosovitskiy2021an} shows that a hybrid approach that generates patch tokens from a CNN model such as ResNet-50 can improve the performance of ViT on the ImageNet1K dataset. Here we experiment with a similar idea by substituting the linear patch embedding in ViT by three convolutional layers as the patch tokenizer. These models are differentiated from others by a suffix $\dagger$ in Table~\ref{table:models}. %For instance, \ours-15$\dagger$. 

\begin{table}[t]
    \centering
    \begin{adjustbox}{max width=\linewidth}
    \begin{tabular}{l|cccccccccc}
        \toprule
                      Model & Patch  & \multicolumn{2}{c}{Patch size} & \multicolumn{2}{c}{Dimension} & \# of heads & $M$ & $r$ \\
                      & embedding & Small & Large & Small & Large &  \\
        \midrule
        \ours-Ti & Linear & 12 & 16 & 96& 192 & 3 & 4 & 4  \\
        \ours-S & Linear & 12 & 16 & 192& 384 & 6 & 4 & 4  \\
        \ours-B & Linear & 12 & 16 & 384& 768 & 12 & 4 & 4  \\
        \ours-9 & Linear & 12 & 16 & 128& 256 & 4 & 3 & 3  \\
        \ours-15 & Linear & 12 & 16 & 192& 384 & 6 & 5 & 3  \\
        \ours-18 & Linear & 12 & 16 & 224& 448 & 7 & 6 & 3  \\
        \ours-9$\dagger$ & 3 Conv. & 12 & 16 & 128& 256 & 4 & 3 & 3  \\
        \ours-15$\dagger$ & 3 Conv. & 12 & 16 & 192& 384 & 6 & 5 & 3  \\
        \ours-18$\dagger$ & 3 Conv. & 12 & 16 & 224& 448 & 7 & 6 & 3  \\
        
        \bottomrule
    \end{tabular}
    \end{adjustbox}
    % \vspace{-2mm}
    \caption{\textbf{Model architectures of \ours.} $K=3$, $N=1$, $L=1$ for all models, and number of heads are same for both branches.
    $K$ denotes the number of multi-scale transformer encoders. $M$, $N$ and $L$ denote the number of transformer encoders of the small and large branches and the cross-attention modules in one multi-scale transformer encoder. $r$ is the expanding ratio of feed-forward network ($\mathtt{FFN}$) in the transformer encoder. See Figure~\ref{fig:cross_attention} for details.}
    % The model with suffix $\dagger$ denotes the linear projection of patch embedding is replaced by three convolutional layers.}
    \label{table:models}
\end{table}

\begin{table}[t]
    \centering
    \begin{adjustbox}{max width=\linewidth}
    \begin{tabular}{l|cccc}
        \toprule
             Model   & Top-1 Acc. & FLOPs & Throughput & Params \\ 
                     & (\%) &  (G) & (images/s) & (M) \\
        \midrule
             DeiT-Ti  & 72.2$\qquad\quad$ & 1.3 & 2557 & 5.7 \\
            \ours-Ti & 73.4 (+1.2) & 1.6 & 1668 & 6.9 \\ 
            % DeiT-9 \\
            \ours-9 & 73.9 (+0.5) & 1.8 & 1530 & 8.6 \\
             \ours-9$\dagger$ & \textbf{77.1} (+3.2) & 2.0 & 1463 & 8.8 \\
            %  DeiT-T$\dagger$ \\
             
            %  DeiT-T$\dagger$  &  \\
            %  \ours-T & 72.6 & 1.6 & 6.9 \\ 
             
            %  \ours-T$\dagger$ &  \\ 
             % another one is 73.75
            %  \ours-15-192 & 73.5 & 1.6 & 7.1 \\
            % \midrule
            % DeiT-9 \\
            %  DeiT-9$\dagger$ \\
             
            %  \ours-12-256 & 76.4 & 2.3 & 10.5 \\
        \midrule
            DeiT-S  &  79.8$\qquad\quad$ & 4.6 & 966 & 22.1 \\
            \ours-S &  81.0 (+1.2) & 5.6 & 690 & 26.7 \\
            \ours-15&  81.5 (+0.5) & 5.8 & 640 & 27.4 \\
            \ours-15$\dagger$ & \textbf{82.3} (+0.8) & 6.1 & 626 & 28.2 \\ 
            \midrule
            % DeiT-15 \\
            
        % \midrule
            DeiT-B  &  81.8$\qquad\quad$ & 17.6 & 314 & 86.6 \\
            \ours-B &  82.2 (+0.4) & 21.2 & 239 & 104.7\\
            % \midrule
            % DeiT-18 \\
            \ours-18& 82.5 (+0.3) & 9.0 & 430 & 43.3 \\ 
        % \midrule
        % \midrule
        % DeiT-S$\dagger$ & 81.0 & 4.8 & 22.6 \\
        % \ours-S$\dagger$ & 82.1 & 5.9 & 27.5 \\
        
        %     DeiT-B  &  81.8 & 17.6 & 86.6 \\
        %     \ours-B &  82.2 & 21.2 & 104.7\\
            \ours-18$\dagger$ & \textbf{82.8} (+0.3) & \textbf{9.5} & 418 & 44.3 \\ 
        \bottomrule
    \end{tabular}
    \end{adjustbox}
    % \vspace{-2mm}
    \caption{\textbf{Comparisons with DeiT baseline on ImageNet1K.} The numbers in the bracket show the improvement from each change. See Table~\ref{table:models} for model details.}
    \label{table:baseline}
\end{table}

% \rc{should we put the changes for FLOPs and parameters in Table 2 as well?}
% \rp{I think since we are focusing more on accuracy, we should not highlight changes for FLOPs.}
% Thus, we compare our results to DeiT as we only use ImageNet1K. 
\subsection{Main Results}
\myparagraphfirst{Comparisons with DeiT.}
% \QF{do we have any ViT results on this? trained by our own using imageNet?} \rc{What is the difference from DeiT? without those data augmentation and regluarization?}
DeiT~\cite{DeiT_touvron2020} is a better trained version of ViT, we thus compare our approach with three baseline models introduced in DeiT, i.e., DeiT-Ti,DeiT-S and DeiT-B. It can be seen from Table~\ref{table:baseline} that \ours improves DeiT-Ti, DeiT-S and DeiT-B by 1.2\%, 1.2\% and 0.4\% points respectively when they are used as the primary branch of \ours. This clearly demonstrates that our proposed cross-attention is effective in learning multi-scale transformer features for image recognition.
%with a simple complementary branch and efficient cross-attention module which brings moderate increase in FLOPs and parameters ($\sim$25\%).  
By making a few architectural changes (see Table~\ref{table:models}), \ours further raises the accuracy of the baselines by another 0.3-0.5\% point, with only a small increase in FLOPs and model parameters. Surprisingly, the convolution-based embedding provides a significant performance boost to \ours-9 (+3.2\%) and \ours-15 (+0.8\%). As the number of transformer encoders increases, the effectiveness of convolution layers seems to become weaker, but \ours-18$\dagger$ still gains another 0.3\% improvement over \ours-18. We would like to point out that the work of T2T~\cite{tokenstotoken_yuan2021} concurrently proposes a different approach based on token-to-token transformation to address the limitation of linear patch embedding in vision transformer.

%in our models (\ours-9$\dagger$ and \ours-15$\dagger$) boosts up the accuracy of \ours-9 by 3.2\% and that of \ours-15 by 0.8\%. The considerable benefit offered by the simple convolution-based embedding suggests that  the patch-based linear embedding in ViT might be a weakness to be addressed in future work on vision transformers. %(\QF{should we mention T2T here})?
%\rc{I am thinking that if we mention the T2T here, will the reviewers it is T2T's contributions? But we definitely need to cite them somehow.}
Despite the design of \ours is intended for accuracy, the efficiency is also considered. E.g., 
% we obtain very competitive efficiency while comparing with 
% The design of \ours is intended for not only accuracy but also efficiency.
\ours-9$\dagger$ and \ours-15$\dagger$ incur 30-50\% more FLOPs and parameters than the baselines. However, their accuracy is considerably improved by $\sim$2.5-5\%. On the other hand, \ours-18$\dagger$ reduces the FLOPs and parameters almost by half compared to DeiT-B while still being 1.0\% more accurate. 
% Both \ours-T and \ours-B improve the baseline by 1.2\% points with moderate increases in FLOPs and parameters. On the other hand, the \ours-9, \ours-15 and \ours-18 improve more when slightly modify the network.

% \rc{\ours-8 is worse than \ours-T by 1\% with similiar FLOPs and parameters.}

%  In this case, our model will be more FLOPs and parameters to the baseline but with better accuracy. Later on, we further trade off between the FLOPs and parameters to result in a model is better in all aspects.

\begin{table}[t]
    \centering
    \begin{adjustbox}{max width=\linewidth}
    \begin{tabular}{l|c|c|c}
        \toprule
             Model   & Top-1 Acc. (\%) & FLOPs (G) & Params (M)  \\ 
        \midrule
            Peceiver~\cite{perceiver_jaegle2021} (arXiv, 2021-03) & 76.4 & $-$ & 43.9 \\
            DeiT-S~\cite{DeiT_touvron2020} (arXiv, 2020-12) & 79.8 & 4.6& 22.1 \\
            % DeiT-S$\dagger$ & 81.0 & 4.8 & 22.6 \\
            CentroidViT-S~\cite{centroidvit_wu2021} (arXiv, 2021-02) & 80.9&4.7&22.3 \\
            PVT-S~\cite{PVT_wang2021} (arXiv, 2021-02) &79.8& 3.8&24.5 \\
            PVT-M~\cite{PVT_wang2021} (arXiv, 2021-02) &81.2&6.7&44.2 \\
            % T2T-ViT-14~\cite{tokenstotoken_yuan2021} (arXiv, 2021-01) & 81.7 & 6.1 & 21.5 \\
            T2T-ViT-14~\cite{tokenstotoken_yuan2021} (arXiv, 2021-01) & 80.7 & 6.1$^*$ & 21.5 \\
            TNT-S~\cite{TNT_han2021transformer} (arXiv, 2021-02) & 81.3& 5.2& 23.8 \\
            \ours-15 (Ours)& 81.5 & 5.8& 27.4 \\
            \ours-15$\dagger$ (Ours) & \textbf{82.3} & 6.1 & 28.2 \\ 
            \midrule
            ViT-B@384~\cite{ViT_dosovitskiy2021an} (ICLR, 2021) & 77.9 & 17.6& 86.6 \\
            % ViT-L@384~\cite{ViT_dosovitskiy2021an} & 76.5 &  \\
            DeiT-B~\cite{DeiT_touvron2020} (arXiv, 2020-12)& 81.8& 17.6& 86.6 \\
            PVT-L~\cite{PVT_wang2021} (arXiv, 2021-02) &81.7&9.8&61.4 \\
            % T2T-ViT-19~\cite{tokenstotoken_yuan2021} (arXiv, 2021-01) & 82.4 & 9.8 & 39.0 \\
            % T2T-ViT-24~\cite{tokenstotoken_yuan2021} (arXiv, 2021-01) & 82.6 & 15.0 & 64.1 \\
            T2T-ViT-19~\cite{tokenstotoken_yuan2021} (arXiv, 2021-01) & 81.4 & 9.8$^*$  & 39.0 \\
            T2T-ViT-24~\cite{tokenstotoken_yuan2021} (arXiv, 2021-01) & 82.2 & 15.0$^*$  & 64.1 \\
            TNT-B~\cite{TNT_han2021transformer} (arXiv, 2021-02) & \textbf{82.8} & 14.1& 65.6 \\
            \ours-18 (Ours)& 82.5& 9.0 & 43.3 \\ 
            % \ours-21& 82.6& 13.3& 63.9 \\
            % \midrule
            % \midrule
            
            % \ours-S$\dagger$ & 82.1 & 5.9 & 27.5 \\
            
            \ours-18$\dagger$ (Ours)& \textbf{82.8} & 9.5 & 44.3 \\ 
            % \ours-15 + T2T & 82.3 & 6.3 & 27.8 \\ 
            % \ours-18 + T2T & 83.0 & 9.5 & 43.4 \\ 
        \bottomrule
        \multicolumn{4}{l}{\footnotesize $^*$: We recompute the flops by using our tools.}
    \end{tabular}
    \end{adjustbox}
    % \vspace{-2mm}
    \caption{\textbf{Comparisons with recent transformer-based models on ImageNet1K.} All models are trained using only ImageNet1K dataset. Numbers are referenced from their recent version as of the submission date.}
    \label{table:sota}
\end{table}
% \QF{DeiT-S$\dagger$ should have more parameters than DeiT-S?} \rc{Fixed}

\vspace{1mm}
\myparagraph{Comparisons with SOTA Transformers.}
% \noindent \textbf{Comparisons with SotA Methods.}
We further compare our proposed approach with some very recent concurrent works on vision transformers. They all improve the original ViT~\cite{ViT_dosovitskiy2021an} with respect to efficiency, accuracy or both. 
% Note that all of them are newly arxived and not published yet by the time our paper was submitted. %\footnote{As of the submission date, only ViT~\cite{ViT_dosovitskiy2021an} is accepted in ICLR 2021 and all others do not appeared in any proceeding.}
As shown in Table~\ref{table:sota}, \ours-15$\dagger$ outperforms the small models of all the other approaches with comparable FLOPs and parameters. 
Interestingly when compared with ViT-B, \ours-18$\dagger$ significantly outperforms it by 4.9\% (77.9\% vs 82.8\%) in accuracy while requiring 50\% less FLOPs and parameters. 
%In terms of FLOPs and parameters, our models are comparable to most of the other approaches.
Furthermore, \ours-18$\dagger$ performs as well as TNT-B and better than the others, but also has fewer FLOPs and parameters. Our approach is consistently better than T2T-ViT~\cite{tokenstotoken_yuan2021} and PVT~\cite{PVT_wang2021} in terms of accuracy and FLOPs, showing the efficacy of multi-scale features in vision transformers.  

\begin{table}[t]
    \centering
    \begin{adjustbox}{max width=\linewidth}
    \begin{tabular}{l|c|c|c|c}
        \toprule
             Model   & Top-1 Acc. & FLOPs & Throughput & Params \\ 
                     & (\%) &  (G) & (images/s) & (M) \\
        \midrule
        
ResNet-101~\cite{ResNet_He_2016_CVPR} & 76.7 & 7.80 & 678 & 44.6 \\
ResNet-152~\cite{ResNet_He_2016_CVPR} & 77.0 & 11.5 & 445 & 60.2 \\
\midrule
ResNeXt-101-32$\times$4d~\cite{ResNeXt_Xie_2017_CVPR} & 78.8 & 8.0 & 477 & 44.2 \\
ResNeXt-101-64$\times$4d~\cite{ResNeXt_Xie_2017_CVPR} & 79.6 & 15.5 & 289 & 83.5 \\
% resnet101@Val: 641.34 Images/second.
% resnet152@Val: 468.36 Images/second.
% resnext50_32x4d@Val: 853.76 Images/second.
% resnext101_32x8d@Val: 290.36 Images/second.
\midrule
SEResNet-101~\cite{SENet_Hu_2018} & 77.6 & 7.8 & 564 & 49.3 \\
SEResNet-152~\cite{SENet_Hu_2018} & 78.4 & 11.5 & 392 & 66.8 \\
SENet-154~\cite{SENet_Hu_2018} & 81.3 & 20.7 & 201 & 115.1\\
\midrule
ECA-Net101~\cite{ECA_wang2020} & 78.7 & 7.4 & 591 & 42.5 \\
ECA-Net152~\cite{ECA_wang2020} & 78.9 & 10.9 & 428 & 59.1 \\
\midrule
RegNetY-8GF~\cite{RegNet_Radosavovic_2020_CVPR} & 79.9 & 8.0 & 557 & 39.2 \\ % && 20.1±0.09
RegNetY-12GF~\cite{RegNet_Radosavovic_2020_CVPR} & 80.3 & 12.1 & 439 & 51.8 \\ %& 19.7±0.06
RegNetY-16GF~\cite{RegNet_Radosavovic_2020_CVPR} & 80.4 & 15.9 & 336 & 83.6 \\
RegNetY-32GF~\cite{RegNet_Radosavovic_2020_CVPR} & 81.0 & 32.3 & 208 & 145.0  \\

\midrule
% EfficienetNet-B0& 77.1& 0.39& 5.3 \\
% EfficienetNet-B1& 79.1& 0.7& 7.8 \\
% EfficienetNet-B2& 80.1& 1.0& 9.2 \\
% EfficienetNet-B3& 81.6& 1.8& 12 \\
EfficienetNet-B4@380~\cite{efficientnet_pmlr_tan_19}& 82.9& 4.2 & 356 & 19 \\
EfficienetNet-B5@456~\cite{efficientnet_pmlr_tan_19}& 83.7& 9.9 & 169 & 30 \\
EfficienetNet-B6@528~\cite{efficientnet_pmlr_tan_19}& 84.0& 19.0 & 100 & 43 \\
EfficienetNet-B7@600~\cite{efficientnet_pmlr_tan_19}& 84.3& 37.0 & 55 & 66 \\
% EfficienetNet-B7+RA~\cite{efficientnet_pmlr_tan_19}& 84.7& 37.0 & 55.09 & 66 \\

% NFNet-F0& 83.6& 12.38& 71.5 \\
% NFNet-F1& 84.7& 35.54& 132.6 \\
% NFNet-F2& 85.1& 62.59& 193.8 \\
% NFNet-F3& 85.7& 114.76& 254.9 \\
% NFNet-F4& 85.9& 215.24& 316.1 \\
% NFNet-F5& 86.0& 289.76& 377.2 \\
            % \midrule
            % BoTNet-T3 & 81.7 & 7.3 & $-$ & 33.5 \\
            % BoTNet-T4 & 82.8 & 10.9 & $-$ & 54.7  \\
            % BoTNet-T5 & 83.5 & 19.3 & $-$ & 75.1 \\
            \midrule
            \ours-15& 81.5 & 5.8 & 640 & 27.4 \\
            \ours-15$\dagger$ & 82.3 & 6.1 & 626 & 28.2 \\ 
            \ours-15$\dagger$@384 & 83.5 & 21.4 & 158 & 28.5 \\ 
            % \ours-18 + T2T & 83.0 & 9.5 & 359.0 & 43.5 \\ 
            \ours-18& 82.5& 9.03 & 430 & 43.3 \\ 
            \ours-18$\dagger$ & 82.8 & 9.5 & 418 & 44.3 \\ 
            % \ours-18@384& 83.7 & 31.1 & 113.70 & 43.3 \\ 
            \ours-18$\dagger$@384& 83.9 & 32.4 & 112 & 44.6 \\ 
            \ours-18$\dagger$@480& 84.1 & 56.6 & 57 & 44.9 \\
            % \ours-18$\dagger$@592& 84.2 & 99.9 & 30.6 & 45.3 \\
            % \midrule
            % \ours-18 + T2T & 83.0 & 12.46 & 43.37 \\
        \bottomrule
    \end{tabular}
    \end{adjustbox}
    % \vspace{-2mm}
    \caption{\textbf{Comparisons with CNN models on ImageNet1K.} Models are evaluated under 224$\times$224 if not specified. 
    The inference throughput is measured under a batch size of 64 on a Nvidia Tesla V100 GPU with cudnn 8.0. We report the averaged speed over 100 iterations.}
    \label{table:compare_cnn}
\end{table}

% deit_base_patch16_224@Val: 308.74 Images/second.
% deit_small_patch16_224@Val: 949.05 Images/second.

% vit_embv7_ca_small_patch12_16_cfg_4_h_224@Val: 640.29 Images/second.

% efficientnet_b7@Val: 55.23 Images/second.
% efficientnet_b6@Val: 97.77 Images/second.
% efficientnet_b5@Val: 168.68 Images/second.
% efficientnet_b4@Val: 356.41 Images/second.
% efficientnet_b3@Val: 731.25 Images/second.

\myparagraph{Comparisons with CNN-based Models.}
CNN-based models are dominant in computer vision applications. In this experiment, we compare our proposed approach with some of the best CNN models including both hand-crafted (e.g., ResNet~\cite{ResNet_He_2016_CVPR}) and search based ones (e.g., EfficientNet~\cite{efficientnet_pmlr_tan_19}).
In addition to accuracy, FLOPs and parameters, run-time speed is measured for all the models and shown as inference throughput (images/second) in Table~\ref{table:compare_cnn}.
We follow prior work~\cite{DeiT_touvron2020} to report accuracy from the original papers.
%In order to compare the CNN model with high-resolution input, we also finetune our model with larger input e.g, 384$\times$384.
First, when compared to the ResNet family, including ResNet~\cite{ResNet_He_2016_CVPR}, ResNeXt~\cite{ResNeXt_Xie_2017_CVPR}, SENet~\cite{SENet_Hu_2018}, ECA-ResNet~\cite{ECA_wang2020} and RegNet~\cite{RegNet_Radosavovic_2020_CVPR}, \ours-15 outperforms all of them in accuracy while being smaller and running more efficiently (except ResNet-101, which is slightly faster).
In addition, our best models such as \ours-15$\dagger$ and \ours-18$\dagger$, when evaluated at higher image resolution, are encouragingly competitive against EfficientNet~\cite{efficientnet_pmlr_tan_19} with regard to accuracy, throughput and parameters. We expect neural architecture search (NAS)~\cite{Nasnet_Zoph_2018_CVPR} to close the performance gap between our approach and EfficientNet. 

% EfficientNet is very efficient in terms of FLOPs and parameters; however, thanks to the simple design in ViT, even though with higher FLOPs counts, our model is much faster than EfficientNet. .... \rc{B5 seems be better in all aspects.}

\vspace{1mm}
\myparagraph{Transfer Learning.}
Despite our model achieves better accuracy on ImageNet1K compared to the baselines (Table~\ref{table:baseline}), it is crucial to check generalization of the models by evaluating transfer performance on tasks with fewer samples.
We validate this by performing transfer learning on 5 image classification tasks, including CIFAR10~\cite{cifar_krizhevsky2009learning}, CIFAR100~\cite{cifar_krizhevsky2009learning}, Pet~\cite{pet_parkhi12a}, CropDisease~\cite{cropdisease_mohanty2016}, and ChestXRay8~\cite{chestxray8_wang2017chestx}. While the first four datasets contains natural images, ChestXRay8 consists of medical images.
We finetune the whole pretrained models with 1,000 epochs, batch size 768, learning rate 0.01, SGD optimizer, weight decay 0.0001, and using the same data augmentation in training on ImageNet1K.
Table~\ref{table:transfer} shows the results. While being better in ImageNet1K, our model is on par with DeiT models on all the downstream classification tasks. This result assures that our models still have good generalization ability rather than only fit to ImageNet1K.

\begin{table}
\centering
\begin{adjustbox}{max width=\linewidth}
\begin{tabular}{l|ccccc}
\toprule
Model &  CIFAR10 &  CIFAR100 &   Pet &  CropDiseases &  ChestXRay8 \\
\midrule
DeiT-S~\cite{DeiT_touvron2020}   &    99.15 &     90.89 & 94.93 &         99.96 &       55.39 \\
DeiT-B~\cite{DeiT_touvron2020}    &    99.10$^*$ & 90.80$^*$  & 94.39 &         99.96 &       55.77 \\  % 98.84 &     89.50
\ours-15 &    99.00 &     90.77 & 94.55 &         99.97 &       55.89 \\
\ours-18  &    99.11 &     91.36 & 95.07 &         99.97 &       55.94 \\
\bottomrule
\multicolumn{6}{l}{\footnotesize $^*$: numbers reported in the original paper.}
\end{tabular}
\end{adjustbox}
\caption{\textbf{Transfer learning performance.} Our \ours models are very competitive with the recent DeiT~\cite{DeiT_touvron2020} models on all the downstream classification tasks.}
\label{table:transfer}
\end{table}
\subsection{Ablation Studies}
\label{subsec:ablations}
In this section, we first compare the different fusion approaches (Section~\ref{subsec:ms_fusion}), and then analyze the effects of different parameters of our architecture design, including the patch sizes, the channel width and depth of the small branch and number of cross-attention modules. At the end, we also validate that the proposed can cooperate with other concurrent works for better accuracy.

% , convention fusion, and as two branches can be considered as two networks, we also experimented with deep mutual learning~\cite{DML_Zhang_2018_CVPR} and on-the-fly knowledge distillation~\cite{ONE_lan2018knowledge} to check the results.

\myparagraph{Comparison of Different Fusion Schemes.} Table~\ref{table:ablation_fusion} shows the performance of different fusions schemes, including (I) no fusion, (II) all-attention, (III) class token fusion, (IV) pairwise fusion, and (V) the proposed cross-attention fusion. Among all the compared strategies, the proposed cross-attention fusion achieves the best accuracy with minor increase in FLOPs and parameters. Surprisingly, despite the use of additional self-attention to combine information between two branches, all-attention fails to achieve better performance compared to the simple class token fusion. While the primary L-branch dominates in accuracy by diminishing the effect of complementary S-branch in other fusion strategies, both of the branches in our proposed cross-attention fusion scheme achieve certain accuracy and their ensemble becomes the best, suggesting that these two branches learn different features for different images.  
%Furthermore, in order to understand whether or not two branches learn different or complementary features, we validate the models by only using \clstoken from one of branches. Interestingly, both branches in the model with cross-attention achieve certain accuracy and the ensemble become the best, this suggests that these two branches learn different features for different images. Unlike all other methods, the complementary branch has very poor accuracy and the performance is fully dominated by the primary branch, which indicates that the complementary branch does not help too much. 

\begin{table}[t]
    \centering
    \begin{adjustbox}{max width=\linewidth}
    \begin{tabular}{l|c|c|c||c|c}
        \toprule
               & Top-1 & FLOPs & Params & \multicolumn{2}{c}{Single Branch Acc. (\%) } \\ 
        Fusion & Acc. (\%)  & (G) & (M) & L-Branch & S-Branch \\
        \midrule
            None & 80.2 & 5.3 & 23.7 & 80.2 & 0.1 \\
            All-Attention & 80.0 &  7.6 & 27.7 & 79.9 & 0.5 \\
            Class Token & 80.3 & 5.4 & 24.2 & 80.6 & 7.6 \\
            Pairwise & 80.3 & 5.5 & 24.2 & 80.3 & 7.3 \\ 
            Cross-Attention & 81.0 & 5.6 & 26.7 & 68.1 & 47.2 \\
        \bottomrule
    \end{tabular}
    \end{adjustbox}
    % \vspace{-2mm}
    \caption{\textbf{Ablation study with different fusions on ImageNet1K.} All models are based on \ours-S. Single branch Acc. is computed using \clstoken from one branch only.}
    \label{table:ablation_fusion}
\end{table}

% \begin{table}[t]
%     \centering
%     \begin{adjustbox}{max width=\linewidth}
%     \begin{tabular}{ccccccc|c|c|c}
%         \toprule
%               Configures & Top-1 & FLOPs & Params \\ 
%                       Fusion & Acc. (\%)  & (G) & (M) \\
%         \midrule
%             None & 80.2 & 5.3 & 23.7 \\ % 0.1\% & 80.2\%
%             Add-all & 80.3 & 5.5 & 24.2 \\ % 
%             Add-Cls & 80.3 & 5.4 & 24.2 \\ % 7.56% & 80.6\%
%             CA & 81.0 & 5.6 & 26.7\\ % 47.2% & 68.1%
%         \bottomrule
%     \end{tabular}
%     \end{adjustbox}
%     % \vspace{-2mm}
%     \caption{Ablation study with different fusions. All models are based on \ours-S.}
%     \label{table:ablation_fusion}
% \end{table}

\myparagraph{Effect of Patch Sizes.} 
%We explore that what is the better patch size pair for \ours. Using to small patch size results in more computational load but it supposes to provide better accuracy.
%Ideally use of small patch sizes 
We perform experiments to understand the effect of patch sizes in our \ours by testing two pairs of patch sizes such as (8, 16) and (12, 16), and observe that the one with (12, 16) achieves better accuracy with fewer FLOPs as shown in Table~\ref{table:ablation} (A). 
Intuitively, (8, 16) should get better results as patch size of 8 provides more fine-grained features; however, it is not good as (12, 16) because of the large difference in granularity between the two branches, which makes it difficult for smooth learning of the features. For the pair (8, 16), the number of patch tokens are 4$\times$ difference while the ratio of patch tokens are only 2$\times$ for the model with (12, 16).

\myparagraph{Channel Width and Depth in S-branch.} Despite our cross-attention is designed to be light-weight, we check the performance by using a more complex S-branch, as shown in Table~\ref{table:ablation} (B and C). Both models increase FLOPs and parameters without any improvement in accuracy, which we think is due to the fact that L-branch has the main role to extract features while S-branch only provides additional information; thus, a light-weight branch is enough.

\myparagraph{Depth of Cross-Attention and Number of Multi-Scale Transformer Encoders.} To increase frequency of fusion across two branches, we can either stack more cross-attention modules ($L$) or stack more multi-scale transformer encoders ($K$) (by reducing $M$ to keep the same total depth of a model). Results are shown in Table~\ref{table:ablation} (D and E). With \ours-S as baseline, too frequent fusion of branches does not provide any performance improvement but introduces more FLOPs and parameters. This is because patch token from the other branch is untouched, and the advantages from stacking more than one cross-attention is small as cross-attention is a linear operation without any nonlinearity function. Likewise, using more multi-scale transformer encoders also does not help in performance which is the similar case to increase the capacity of S-branch.

\myparagraph{Importance of CLS Tokens.} We experiment with one model based on \ours-S without CLS tokens, where the model averages the patch tokens of one branch as the CLS token for cross attention with the other branch.
This model achieved 80.0\% accuracy which is is 1\% worse than \ours-S (81.0\%) on ImageNet1K, showing effectiveness of CLS token in summarizing information of current branch for passing to another one through cross-attention. 

\myparagraph{Cooperation with Concurrent Works.} Our proposed cross-attention is also capable of cooperating with other concurrent ViT variants. We consider T2T-ViT~\cite{tokenstotoken_yuan2021} as a case study and use the T2T module to replace linear projection of patch embedding in both branches on \ours-18. \ours-18+T2T achieves an top-1 accuracy of 83.0\% on ImageNet1K, additional 0.5\% improvement over \ours-18. This shows that our proposed cross-attention is also capable of learning multi-scale features for other ViT variants. 

% Additional results and discussions are included in the supplementary material.

\begin{table}[t]
    \centering
    \begin{adjustbox}{max width=\linewidth}
    \begin{tabular}{l|cccccccc|c|c|c}
        \toprule
        Model        & \multicolumn{2}{c}{Patch size} & \multicolumn{2}{c}{Dimension} & & & & & Top-1 & FLOPs & Params \\ 
                & Small     & Large               &   Small       &   Large & $K$ & $N$ & $M$ & $L$ & Acc. (\%)  & (G) & (M) \\
        \midrule
        \ours-S &    12 & 16 & 192& 384 & 3 & 1 & 4 & 1 & 81.0 & 5.6 & 26.7\\
        \midrule
        A &    \boldblue{8}  & 16 & 192 & 384 & 3 & 1 & 4 & 1 & 80.8 & 6.7 & 26.7\\
        % A &    8  & \textcolor{gray}{16} & \textcolor{gray}{192} & \textcolor{gray}{384} & \textcolor{gray}{3} & \textcolor{gray}{1} & \textcolor{gray}{4} & \textcolor{gray}{1} & 80.8 & 6.7 & 26.7\\
        B &    12 & 16 & \boldblue{384} & 384 & 3 & 1 & 4 & 1 & 80.1 & 7.7 & 31.4 \\
        C &    12 & 16 & 192& 384 & 3 & \boldblue{2} & 4 & 1 & 80.7 & 6.3 & 28.0 \\
        D &    12 & 16 & 192& 384 & 3 & 1 & 4 & \boldblue{2} & 81.0 & 5.6 & 28.9 \\
        E &    12 & 16 & 192& 384 & \boldblue{6} & 1 & \boldblue{2} & 1 & 80.9 & 6.6 & 31.1\\
        \bottomrule
    \end{tabular}
    \end{adjustbox}
    % \vspace{-2mm}
    \caption{\textbf{Ablation study with different architecture parameters on ImageNet1K.} The \boldblue{blue} color indicates changes from \ours-S.}
    \label{table:ablation}
\end{table}
% \ours-18+T2T: 83.0\%

% \ours-15+T2T: 82.3

% \begin{enumerate}
%     \item The patch size selection. - 8, 16:   12, 16
%     \item Channel width ratio of cheap branch, - 2,  should I try 1, 4, or 1.5? (4 results in very thin net)
%     \item Number of blocks of cheap branch - 1, 2, now, using 1 or 2 blocks do not matter, but those are tested under 4 blocks in the main branch, maybe I can try more when using 6 blocks in the main branch
%     \item - Number of exchanges, cross attention - more frequent exchanges, I tried 3 and 6 now.  6 is not better but no harm (should we discuss the place of ca modules?)
%     \item - Depth of cross attention - usually 1, and trying 2 now (NaN)
%     \item - Other fusion w.r.t. cross attention - add cls token    - combining all patches via upsampling or downsampling
%     \item - Other training setting?, deep mutaul learing and one learning
    
% \end{enumerate}

% \subsection{When using ImageNet21k??}

% \subsection{Others}

% \begin{enumerate}
%     \item feature visualization?
%     % \item transfer learning
% \end{enumerate}
\section{Conclusion}
\label{sec:conclusion}
In this paper, we present \ours, a dual-branch vision transformer for learning multi-scale features, to improve the recognition accuracy for image classification. To effectively combine image patch tokens of different scales, we further develop a fusion method based on cross-attention to exchange information between two branches efficiently in linear time.
%Our \ours processes small-patch and large-patch tokens with two separate branches of different computational complexity to balance the workload and these tokens are then fused purely by attention multiple times to complement each other. 
%Furthermore, we propose an effective cross-attention module for multi-scale feature fusion that uses a single token for each branch as a query to exchange information with other branches. The cross-attention only requires linear time for both computational and memory complexity instead of quadratic time otherwise. 
With extensive experiments, we demonstrate that our proposed model performs better than or on par with several concurrent works on vision transformer, in addition to efficient CNN models.
While our current work scratches the surface on multi-scale vision transformers for image classification, we anticipate that in future there will be more works in developing efficient multi-scale transformers for other vision applications, including object detection, semantic segmentation, and video action recognition.
% Those results also validate that multi-scale feature representations are applicable for transformer-based models, we anticipate that there will more works on cooperating vision knowledge into vision transformer.

% The proposed cross-attention module efficiently and effectively improve the baseline models by 1.2\% for the smaller models with $\sim$25\% increase on FLOPs and parameters; furthermore, with slight modification on the architecture of the baseline models, e.g., changing depth and width of a model and replacing linear projection with three convolutions, \ours can further outperform the baselines by a significant amount. 
% \ours also outperform many concurrent works on ViT variants when trading off among accuracy, FLOPs and parameters.
% Moreover, \ours is also on par with search-based CNN models, we anticipate that transformer-based model can be further improved with architecture search.

{\small
\bibliographystyle{ieee_fullname}
\bibliography{egbib}

\begin{thebibliography}{10}\itemsep=-1pt

\bibitem{adelson1984pyramid}
Edward~H Adelson, Charles~H Anderson, James~R Bergen, Peter~J Burt, and Joan~M
  Ogden.
\newblock Pyramid methods in image processing.
\newblock {\em RCA engineer}, 29(6):33--41, 1984.

\bibitem{lambdanetworks_bello2021}
Irwan Bello.
\newblock Lambdanetworks: Modeling long-range interactions without attention.
\newblock In {\em International Conference on Learning Representations}, 2021.

\bibitem{bello2019attention}
Irwan Bello, Barret Zoph, Ashish Vaswani, Jonathon Shlens, and Quoc~V Le.
\newblock Attention augmented convolutional networks.
\newblock In {\em Proceedings of the IEEE/CVF International Conference on
  Computer Vision}, pages 3286--3295, 2019.

\bibitem{cai2016unified}
Zhaowei Cai, Quanfu Fan, Rogerio~S Feris, and Nuno Vasconcelos.
\newblock A unified multi-scale deep convolutional neural network for fast
  object detection.
\newblock In {\em European conference on computer vision}, pages 354--370.
  Springer, 2016.

\bibitem{chen2018big}
Chun-Fu~(Richard) Chen, Quanfu Fan, Neil Mallinar, Tom Sercu, and Rogerio
  Feris.
\newblock {Big-Little Net: An Efficient Multi-Scale Feature Representation for
  Visual and Speech Recognition}.
\newblock In {\em International Conference on Learning Representations}, 2019.

\bibitem{chen2019drop}
Yunpeng Chen, Haoqi Fan, Bing Xu, Zhicheng Yan, Yannis Kalantidis, Marcus
  Rohrbach, Shuicheng Yan, and Jiashi Feng.
\newblock Drop an octave: Reducing spatial redundancy in convolutional neural
  networks with octave convolution.
\newblock In {\em Proceedings of the IEEE/CVF International Conference on
  Computer Vision}, pages 3435--3444, 2019.

\bibitem{HigherHRNet_Cheng_2020_CVPR}
Bowen Cheng, Bin Xiao, Jingdong Wang, Honghui Shi, Thomas~S. Huang, and Lei
  Zhang.
\newblock Higherhrnet: Scale-aware representation learning for bottom-up human
  pose estimation.
\newblock In {\em IEEE/CVF Conference on Computer Vision and Pattern
  Recognition}, June 2020.

\bibitem{RandAug_NEURIPS2020_Cubuk}
Ekin~Dogus Cubuk, Barret Zoph, Jon Shlens, and Quoc Le.
\newblock {RandAugment: Practical Automated Data Augmentation with a Reduced
  Search Space}.
\newblock In H Larochelle, M Ranzato, R Hadsell, M~F Balcan, and H Lin,
  editors, {\em Advances in Neural Information Processing Systems}, pages
  18613--18624. Curran Associates, Inc., 2020.

\bibitem{imagenet_deng2009}
Jia Deng, Wei Dong, Richard Socher, Li-Jia Li, Kai Li, and Li Fei-Fei.
\newblock Imagenet: A large-scale hierarchical image database.
\newblock In {\em 2009 IEEE conference on computer vision and pattern
  recognition}, pages 248--255. Ieee, 2009.

\bibitem{devlin2018bert}
Jacob Devlin, Ming-Wei Chang, Kenton Lee, and Kristina Toutanova.
\newblock {BERT}: Pre-training of deep bidirectional transformers for language
  understanding.
\newblock In {\em Proceedings of the 2019 Conference of the North {A}merican
  Chapter of the Association for Computational Linguistics: Human Language
  Technologies, Volume 1 (Long and Short Papers)}, pages 4171--4186,
  Minneapolis, Minnesota, June 2019. Association for Computational Linguistics.

\bibitem{ViT_dosovitskiy2021an}
Alexey Dosovitskiy, Lucas Beyer, Alexander Kolesnikov, Dirk Weissenborn,
  Xiaohua Zhai, Thomas Unterthiner, Mostafa Dehghani, Matthias Minderer, Georg
  Heigold, Sylvain Gelly, Jakob Uszkoreit, and Neil Houlsby.
\newblock An image is worth 16x16 words: Transformers for image recognition at
  scale.
\newblock In {\em International Conference on Learning Representations}, 2021.

\bibitem{fan2019more}
Quanfu Fan, Chun-Fu~Richard Chen, Hilde Kuehne, Marco Pistoia, and David Cox.
\newblock More is less: Learning efficient video representations by big-little
  network and depthwise temporal aggregation.
\newblock In {\em Advances in Neural Information Processing Systems}, pages
  2261--2270, 2019.

\bibitem{feichtenhofer2019slowfast}
Christoph Feichtenhofer, Haoqi Fan, Jitendra Malik, and Kaiming He.
\newblock Slowfast networks for video recognition.
\newblock In {\em Proceedings of the IEEE International Conference on Computer
  Vision}, pages 6202--6211, 2019.

\bibitem{TNT_han2021transformer}
Kai Han, An Xiao, Enhua Wu, Jianyuan Guo, Chunjing Xu, and Yunhe Wang.
\newblock Transformer in transformer.
\newblock {\em arXiv preprint arXiv:2103.00112}, 2021.

\bibitem{ResNet_He_2016_CVPR}
Kaiming He, Xiangyu Zhang, Shaoqing Ren, and Jian Sun.
\newblock {Deep Residual Learning for Image Recognition}.
\newblock In {\em The IEEE Conference on Computer Vision and Pattern
  Recognition}, June 2016.

\bibitem{InstanceRepetition_Hoffer_2020_CVPR}
Elad Hoffer, Tal Ben-Nun, Itay Hubara, Niv Giladi, Torsten Hoefler, and Daniel
  Soudry.
\newblock Augment your batch: Improving generalization through instance
  repetition.
\newblock In {\em Proceedings of the IEEE/CVF Conference on Computer Vision and
  Pattern Recognition}, June 2020.

\bibitem{hu2019local}
Han Hu, Zheng Zhang, Zhenda Xie, and Stephen Lin.
\newblock Local relation networks for image recognition.
\newblock In {\em Proceedings of the IEEE/CVF International Conference on
  Computer Vision}, pages 3464--3473, 2019.

\bibitem{SENet_Hu_2018}
J. {Hu}, L. {Shen}, and G. {Sun}.
\newblock Squeeze-and-excitation networks.
\newblock In {\em 2018 IEEE/CVF Conference on Computer Vision and Pattern
  Recognition}, pages 7132--7141, 2018.

\bibitem{perceiver_jaegle2021}
Andrew Jaegle, Felix Gimeno, Andy Brock, Oriol Vinyals, Andrew Zisserman, and
  Joao Carreira.
\newblock Perceiver: General perception with iterative attention.
\newblock In Marina Meila and Tong Zhang, editors, {\em Proceedings of the 38th
  International Conference on Machine Learning}, volume 139 of {\em Proceedings
  of Machine Learning Research}, pages 4651--4664. PMLR, 18--24 Jul 2021.

\bibitem{cifar_krizhevsky2009learning}
Alex Krizhevsky, Geoffrey Hinton, et~al.
\newblock Learning multiple layers of features from tiny images.
\newblock 2009.

\bibitem{SKNet_Li_2019_CVPR}
Xiang Li, Wenhai Wang, Xiaolin Hu, and Jian Yang.
\newblock Selective kernel networks.
\newblock In {\em Proceedings of the IEEE/CVF Conference on Computer Vision and
  Pattern Recognition}, June 2019.

\bibitem{lin2017feature}
Tsung-Yi Lin, Piotr Doll{\'a}r, Ross Girshick, Kaiming He, Bharath Hariharan,
  and Serge Belongie.
\newblock Feature pyramid networks for object detection.
\newblock In {\em Proceedings of the IEEE conference on computer vision and
  pattern recognition}, pages 2117--2125, 2017.

\bibitem{cropdisease_mohanty2016}
Sharada~P Mohanty, David~P Hughes, and Marcel Salath{\'e}.
\newblock Using deep learning for image-based plant disease detection.
\newblock {\em Frontiers in plant science}, 7:1419, 2016.

\bibitem{Deblurring_Nah_2017_CVPR}
Seungjun Nah, Tae Hyun~Kim, and Kyoung Mu~Lee.
\newblock Deep multi-scale convolutional neural network for dynamic scene
  deblurring.
\newblock In {\em Proceedings of the IEEE Conference on Computer Vision and
  Pattern Recognition}, July 2017.

\bibitem{hourglass}
Alejandro Newell, Kaiyu Yang, and Jia Deng.
\newblock {Stacked Hourglass Networks for Human Pose Estimation}.
\newblock In Bastian Leibe, Jiri Matas, Nicu Sebe, and Max Welling, editors,
  {\em Proceedings of the European Conference on Computer Vision}, pages
  483--499, Cham, 2016. Springer International Publishing.

\bibitem{newell2016stacked}
Alejandro Newell, Kaiyu Yang, and Jia Deng.
\newblock Stacked hourglass networks for human pose estimation.
\newblock In {\em European conference on computer vision}, pages 483--499.
  Springer, 2016.

\bibitem{pet_parkhi12a}
Omkar~M. Parkhi, Andrea Vedaldi, Andrew Zisserman, and C.~V. Jawahar.
\newblock Cats and dogs.
\newblock In {\em IEEE Conference on Computer Vision and Pattern Recognition},
  2012.

\bibitem{pedersoli2015coarse}
Marco Pedersoli, Andrea Vedaldi, Jordi Gonzalez, and Xavier Roca.
\newblock A coarse-to-fine approach for fast deformable object detection.
\newblock {\em Pattern Recognition}, 48(5):1844--1853, 2015.

\bibitem{perona1990scale}
Pietro Perona and Jitendra Malik.
\newblock Scale-space and edge detection using anisotropic diffusion.
\newblock {\em IEEE Transactions on pattern analysis and machine intelligence},
  12(7):629--639, 1990.

\bibitem{RegNet_Radosavovic_2020_CVPR}
Ilija Radosavovic, Raj~Prateek Kosaraju, Ross Girshick, Kaiming He, and Piotr
  Dollar.
\newblock Designing network design spaces.
\newblock In {\em IEEE/CVF Conference on Computer Vision and Pattern
  Recognition}, June 2020.

\bibitem{SASA_Ramachandran_2019_NeurIPS}
Prajit Ramachandran, Niki Parmar, Ashish Vaswani, Irwan Bello, Anselm Levskaya,
  and Jon Shlens.
\newblock {Stand-Alone Self-Attention in Vision Models}.
\newblock In H Wallach, H Larochelle, A Beygelzimer, F~d Alch~e Buc, E Fox, and
  R Garnett, editors, {\em Advances in Neural Information Processing Systems}.
  Curran Associates, Inc., 2019.

\bibitem{BoT_srinivas2021}
Aravind Srinivas, Tsung-Yi Lin, Niki Parmar, Jonathon Shlens, Pieter Abbeel,
  and Ashish Vaswani.
\newblock Bottleneck transformers for visual recognition.
\newblock {\em arXiv preprint arXiv:2101.11605}, 2021.

\bibitem{JFT300M_ICCV_2017}
C. {Sun}, A. {Shrivastava}, S. {Singh}, and A. {Gupta}.
\newblock Revisiting unreasonable effectiveness of data in deep learning era.
\newblock In {\em 2017 IEEE International Conference on Computer Vision}, pages
  843--852, 2017.

\bibitem{efficientnet_pmlr_tan_19}
Mingxing Tan and Quoc Le.
\newblock {EfficientNet: Rethinking Model Scaling for Convolutional Neural
  Networks}.
\newblock In Kamalika Chaudhuri and Ruslan Salakhutdinov, editors, {\em
  Proceedings of the 36th International Conference on Machine Learning}, pages
  6105--6114, Long Beach, California, USA, June 2019. PMLR.

\bibitem{DeiT_touvron2020}
Hugo Touvron, Matthieu Cord, Matthijs Douze, Francisco Massa, Alexandre
  Sablayrolles, and Herve Jegou.
\newblock Training data-efficient image transformers \& distillation through
  attention.
\newblock In Marina Meila and Tong Zhang, editors, {\em Proceedings of the 38th
  International Conference on Machine Learning}, volume 139 of {\em Proceedings
  of Machine Learning Research}, pages 10347--10357. PMLR, 18--24 Jul 2021.

\bibitem{Transformer_NIPS2017_Vaswani}
Ashish Vaswani, Noam Shazeer, Niki Parmar, Jakob Uszkoreit, Llion Jones,
  Aidan~N Gomez, ukasz Kaiser, and Illia Polosukhin.
\newblock {Attention is All you Need}.
\newblock In I Guyon, U~V Luxburg, S Bengio, H Wallach, R Fergus, S
  Vishwanathan, and R Garnett, editors, {\em Advances in Neural Information
  Processing Systems}. Curran Associates, Inc., 2017.

\bibitem{ECA_wang2020}
Qilong Wang, Banggu Wu, Pengfei Zhu, Peihua Li, Wangmeng Zuo, and Qinghua Hu.
\newblock Eca-net: Efficient channel attention for deep convolutional neural
  networks.
\newblock In {\em The IEEE Conference on Computer Vision and Pattern
  Recognition}, 2020.

\bibitem{PVT_wang2021}
Wenhai Wang, Enze Xie, Xiang Li, Deng-Ping Fan, Kaitao Song, Ding Liang, Tong
  Lu, Ping Luo, and Ling Shao.
\newblock Pyramid vision transformer: A versatile backbone for dense prediction
  without convolutions, 2021.

\bibitem{wang2018non}
Xiaolong Wang, Ross Girshick, Abhinav Gupta, and Kaiming He.
\newblock Non-local neural networks.
\newblock In {\em Proceedings of the IEEE conference on computer vision and
  pattern recognition}, pages 7794--7803, 2018.

\bibitem{chestxray8_wang2017chestx}
Xiaosong Wang, Yifan Peng, Le Lu, Zhiyong Lu, Mohammadhadi Bagheri, and
  Ronald~M Summers.
\newblock Chestx-ray8: Hospital-scale chest x-ray database and benchmarks on
  weakly-supervised classification and localization of common thorax diseases.
\newblock In {\em Proceedings of the IEEE conference on computer vision and
  pattern recognition}, pages 2097--2106, 2017.

\bibitem{CBAM_Woo_2018_ECCV}
Sanghyun Woo, Jongchan Park, Joon-Young Lee, and In~So Kweon.
\newblock Cbam: Convolutional block attention module.
\newblock In {\em Proceedings of the European Conference on Computer Vision},
  September 2018.

\bibitem{centroidvit_wu2021}
Lemeng Wu, Xingchao Liu, and Qiang Liu.
\newblock Centroid transformers: Learning to abstract with attention.
\newblock {\em arXiv preprint arXiv:2102.08606}, 2021.

\bibitem{ResNeXt_Xie_2017_CVPR}
Saining Xie, Ross Girshick, Piotr Doll{\'a}r, Zhuowen Tu, and Kaiming He.
\newblock {Aggregated Residual Transformations for Deep Neural Networks}.
\newblock In {\em The IEEE Conference on Computer Vision and Pattern
  Recognition}, July 2017.

\bibitem{yang2015multi}
Songfan Yang and Deva Ramanan.
\newblock Multi-scale recognition with dag-cnns.
\newblock In {\em Proceedings of the IEEE international conference on computer
  vision}, pages 1215--1223, 2015.

\bibitem{tokenstotoken_yuan2021}
Li Yuan, Yunpeng Chen, Tao Wang, Weihao Yu, Yujun Shi, Francis~EH Tay, Jiashi
  Feng, and Shuicheng Yan.
\newblock Tokens-to-token vit: Training vision transformers from scratch on
  imagenet, 2021.

\bibitem{CutMix_Yun_2019_ICCV}
Sangdoo Yun, Dongyoon Han, Seong~Joon Oh, Sanghyuk Chun, Junsuk Choe, and
  Youngjoon Yoo.
\newblock {CutMix: Regularization Strategy to Train Strong Classifiers With
  Localizable Features}.
\newblock In {\em Proceedings of the IEEE/CVF International Conference on
  Computer Vision}, Oct. 2019.

\bibitem{Mixup_zhang2018}
Hongyi Zhang, Moustapha Cisse, Yann~N. Dauphin, and David Lopez-Paz.
\newblock mixup: Beyond empirical risk minimization.
\newblock In {\em International Conference on Learning Representations}, 2018.

\bibitem{SAN_Zhao_2020_CVPR}
Hengshuang Zhao, Jiaya Jia, and Vladlen Koltun.
\newblock Exploring self-attention for image recognition.
\newblock In {\em Proceedings of the IEEE/CVF Conference on Computer Vision and
  Pattern Recognition}, June 2020.

\bibitem{RandomErasing_Zhong_2020}
Zhun Zhong, Liang Zheng, Guoliang Kang, Shaozi Li, and Yi Yang.
\newblock {Random Erasing Data Augmentation}.
\newblock {\em Proceedings of the AAAI Conference on Artificial Intelligence},
  34(07):13001--13008, Apr. 2020.

\bibitem{Nasnet_Zoph_2018_CVPR}
Barret Zoph, Vijay Vasudevan, Jonathon Shlens, and Quoc~V. Le.
\newblock Learning transferable architectures for scalable image recognition.
\newblock In {\em Proceedings of the IEEE Conference on Computer Vision and
  Pattern Recognition}, June 2018.

\end{thebibliography}
}

\clearpage

\appendix
\myparagraphfirst{Summary} This supplementary material contains the following additional comparisons and hyperparameter details.
We first provide more comparisons between the proposed \ours and DeiT (see Table~\ref{table:more_baselines}) and then list the training hyperparameters used in main results, ablation studies and transfer learning, in Table~\ref{table:training_params}.

\section{More Comparisons and Analysis}

% \subimport{./}{Supplemental/Tables/c1}
% \subimport{./}{Supplemental/Tables/c2}

To further check the advantages of the proposed \ours, we trained the models whose architecture are identical to the L-branch (primary) of our models. E.g., DeiT-9 is the baseline for \ours-9. As shown in Table~\ref{table:more_baselines},
the proposed cross-attention fusion consistently improves the baseline vision transformers regardless of their primary branches and patch embeddings, suggesting that the proposed multi-scale fusion is effective for different vision transformers. 
% E.g., \ours-18$\dagger$ outperforms DeiT-18$\dagger$ by 1.6\%.

\begin{table}[!b]
    \centering
    \begin{adjustbox}{max width=\linewidth}
    \begin{tabular}{l|c|c|c}
        \toprule
             Model   & Top-1 Acc. (\%) & FLOPs (G) & Params (M)  \\ 
        \midrule
            %  DeiT-Ti  & 72.2 & 1.3 & 5.7 \\
            % \ours-Ti & 73.4 & 1.6 & 6.9 \\ 
            % DeiT-Ti$\dagger$ & 75.3 &  1.3 & 5.8 \\
            % DeiT-Ti$\dagger$ & 74.4 &  1.3 & 5.8 \\
            % \ours-Ti$\dagger$ & 74.7 & 1.6 & 7.1 \\ 
            DeiT-9 & 72.9 & 1.4 & 6.4\\
            \ours-9 & 73.9 & 1.8 & 8.6 \\
            DeiT-9$\dagger$ & 75.6 &  1.5 & 6.6\\
             \ours-9$\dagger$ & \textbf{77.1} & 2.0 & 8.8 \\
             
            %  DeiT-T$\dagger$  &  \\
            %  \ours-T & 72.6 & 1.6 & 6.9 \\ 

             % another one is 73.75
            %  \ours-15-192 & 73.5 & 1.6 & 7.1 \\
            % \midrule
            % DeiT-9 \\
            %  DeiT-9$\dagger$ \\
             
            %  \ours-12-256 & 76.4 & 2.3 & 10.5 \\
        \midrule
            % DeiT-S  &  79.8 & 4.6 & 22.1 \\
            % \ours-S &  81.0 & 5.6 & 26.7 \\
            % DeiT-S$\dagger$  &  81.6 & 4.8 & 22.6 \\
            % \ours-S$\dagger$ &  82.1 & 5.6 & 26.7 \\
            DeiT-15 & 80.8 & 4.9 & 22.9 \\
            \ours-15&  81.5 & 5.8& 27.4 \\
            DeiT-15$\dagger$ & 81.7 & 5.1 & 23.5\\
            \ours-15$\dagger$ & \textbf{82.3} & 6.1 & 28.2 \\ 
            \midrule
            % DeiT-15 \\
        % \midrule
            % DeiT-B  &  81.8 & 17.6 & 86.6 \\
            % \ours-B &  82.2 & 21.2 & 104.7\\
            % \midrule
            DeiT-18 & 81.4 & 7.8 & 37.1\\
            \ours-18& 82.5 & 9.0 & 43.3 \\ 
        % \midrule
        % \midrule
        % DeiT-S$\dagger$ & 81.0 & 4.8 & 22.6 \\
        % \ours-S$\dagger$ & 82.1 & 5.9 & 27.5 \\
        
        %     DeiT-B  &  81.8 & 17.6 & 86.6 \\
        %     \ours-B &  82.2 & 21.2 & 104.7\\
        DeiT-18$\dagger$ & 81.2 & 8.1 & 37.9 \\
            \ours-18$\dagger$ & \textbf{82.8} & 9.5 & 44.3 \\ 
        \bottomrule
    \end{tabular}
    \end{adjustbox}
    % \vspace{-2mm}
    \caption{\textbf{Comparisons with various baselines on ImageNet1K.} See Table~1 of the main paper for model details. $\dagger$ denotes the models using three convolutional layers for patch embedding instead of linear projection.}
    \label{table:more_baselines}
\end{table}

% \begin{table*}[bh!]
%     \centering
%     \begin{adjustbox}{max width=\linewidth}
%     \begin{tabular}{cccccccccccccc}
%         \toprule
%         & Batch size & Epochs & Optimizer & Weight & Linear-rate Scheduler & Warmup & Warmup linear-rate & Data Aug. & Mixup  & CutMix ($\alpha$) & Random  & Instance & Drop path \\
%         &             &       &           & Decay  &  (Initial LR) &         Epochs &     scheduler (Initial LR)              &           &  ($\alpha$) & ($\alpha$)    &  Erasing &  Repetition$^*$   \\                                                 
%         \midrule
%         Main Results & 4,096 & 300 & AdamW & 0.05 & Cosine (0.004) & 30 & Linear (1e-6) & RandAugment (m=9, n=2) & 0.8 & 1.0 & 0.25 & 3 & 0.1 \\ 
%         \midrule
%         Transfer & 768 & 1,000 & SGD (0.9) & 1e-4 & Cosine (0.01) & 5 & Linear (1e-6) & RandAugment (m=9, n=2) & 0.8 & 1.0 & 0.0 & 3 & 0.0 \\ 
%         \bottomrule
%         \multicolumn{13}{l}{\footnotesize $^*$: only used for \ours-18.}
%     \end{tabular}
%     \end{adjustbox}
%     % \vspace{-2mm}
%     \caption{\textbf{Details of training settings.}}
%     \label{table:training_params}
% \end{table*}

\begin{table}[!bt]
    \centering
    \begin{adjustbox}{max width=\linewidth}
    \begin{tabular}{c|cc}
        \toprule
            & Main Results & Transfer \\
        \midrule
        Batch size & 4,096 & 768 \\
        Epochs & 300 & 1,000 \\
        Optimizer & AdamW & SGD \\
        Weight Decay & 0.05 & 1e-4 \\
        Linear-rate Scheduler & \multirow{2}{*}{Cosine (0.004)} & \multirow{2}{*}{Cosine (0.01)} \\
        (Initial LR) \\
        \midrule
        Warmup Epochs & 30 & 5 \\
        Warmup linear-rate & \multicolumn{2}{c}{\multirow{2}{*}{Linear (1e-6)}} \\
        Scheduler (Initial LR) \\
        \midrule
        Data Aug. & \multicolumn{2}{c}{RandAugment (m=9, n=2)} \\
        \midrule
        Mixup ($\alpha$) & \multicolumn{2}{c}{0.8} \\
        CutMix ($\alpha$) & \multicolumn{2}{c}{1.0} \\
        Random Erasing & 0.25 & 0.0 \\
        \midrule
        Instance & \multicolumn{2}{c}{\multirow{2}{*}{3}} \\
        Repetition$^*$ \\
        \midrule
        Drop-path & 0.1 & 0.0 \\
        Label Smoothing & \multicolumn{2}{c}{0.1} \\
        \bottomrule
        \multicolumn{3}{l}{\footnotesize $^*$: only used for \ours-18.}
    \end{tabular}
    \end{adjustbox}
    % \vspace{-2mm}
    \caption{\textbf{Details of training settings.}}
    \label{table:training_params}
\end{table}

\begin{figure*}[h]
    \centering
    \includegraphics[width=\linewidth]{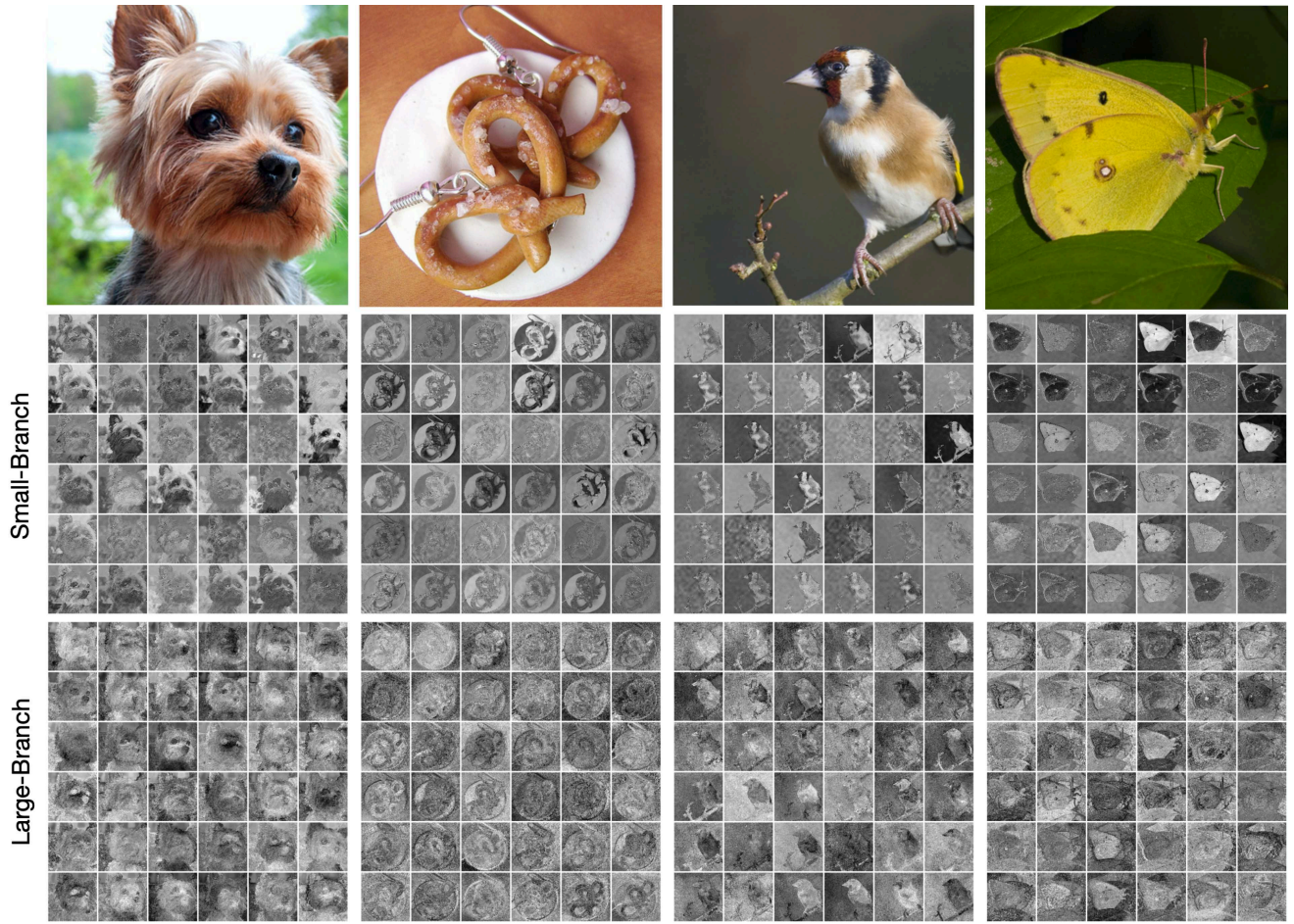}
    \caption{\textbf{Feature visualization of \ours-S.} Features of patch tokens of both branches from the last multi-scale transformer encoder are shown. (36 random channels are selected.)
    }
    \label{fig:feat} 
    % \vspace{-3mm}
\end{figure*}

Figure~\ref{fig:feat} visualizes the features of both branches from the last multi-scale transformer encoder of \ours. 
The proposed cross-attention learns different features in both branches, where the small branch generates more low-level features because there are only three transformer encoders while the features of the large branch are more abstract.  Both branches complement each other and hence the ensemble results are better.

\end{document}